# Hardware-Software Collaborative Computing of Photonic Spiking Reinforcement Learning for Robotic Continuous Control

Mengting Yu[1], Shuiying Xiang[1*], Changjian Xie[1*], Yonghang Chen[1], Haowen Zhao[1], Xingxing Guo[1], Yahui Zhang[1], Yanan Han[1], & Yue Hao[1]

[1] State Key Laboratory of Integrated Service Networks, State Key Discipline Laboratory of Wide Bandgap Semiconductor Technology, Xidian University, Xi'an 710071, China.

**Abstract:** Robotic continuous control tasks impose stringent demands on the energy efficiency and latency of computing architectures due to their high-dimensional state spaces and real-time interaction requirements. Conventional electronic computing platforms face computational bottlenecks, whereas the fusion of photonic computing and spiking reinforcement learning (RL) offers a promising alternative. Here, we propose a novel computing architecture based on photonic spiking RL, which integrates the Twin Delayed Deep Deterministic policy gradient (TD3) algorithm with spiking neural network (SNN). The proposed architecture employs an optical-electronic hybrid computing paradigm wherein a silicon photonic Mach-Zehnder interferometer (MZI) chip executes linear matrix computations, while nonlinear spiking activations are performed in the electronic domain. Experimental validation on the Pendulum-v1 and HalfCheetah-v2 benchmarks demonstrates the system's capability for software-hardware co-inference, achieving a control policy reward of $5831 \pm 495$ on HalfCheetah-v2, a 23.33% reduction in convergence steps, and an action deviation below 2.2%. Notably, this work represents the first application of a programmable MZI photonic computing chip to robotic continuous control tasks, attaining an energy efficiency of 1.39 TOPS/W and an ultralow computational latency of 120 ps. Such performance underscores the promise of photonic spiking RL for real-time decision-making in autonomous and industrial robotic systems.

Address all correspondence to Shuiying Xiang, email: syxiang@xidian.edu.cn; Changjian Xie, email: xiechangjian@xidian.edu.cn.

## 1. Introduction

The rapid proliferation of robotics in industrial manufacturing, social services, and healthcare has imposed increasingly stringent demands on environmental perception, autonomous decision-making, and motion control [1]. Advances in artificial intelligence (AI), particularly in deep reinforcement learning (DRL), have opened new pathways for developing intelligent robotic systems with enhanced adaptive capabilities [2]-[6]. Currently, DRL-based robotic continuous control has emerged as a central research focus at the intersection of robotics and artificial intelligence. It has achieved remarkable progress in representative tasks such as autonomous navigation [7]-[10], object grasping [11]-[14], and gait control [15]-[18], and these advances can be further extended to more complex applications including emergency rescue and deep space exploration. However, as the critical link between intelligent decision-making and physical execution, robotic continuous control still faces three core challenges: (1) efficient modeling of complex dynamics in high-dimensional state spaces, (2) real-time computation under constraints of limited computational power and energy efficiency, and (3) maintaining system stability in highly latency-sensitive control processes [19]-[22]. These challenges are particularly pronounced in continuous control tasks on simulation platforms like MuJoCo, highlighting the urgent need for innovative solutions.

To address the challenges of continuous action modeling in high-dimensional state spaces, researchers have proposed a series of DRL-based solutions for typical continuous control tasks on simulation platforms such as MuJoCo. Lillicrap et al. [19] and Silver et al. [23] introduced the Deep Deterministic Policy Gradient (DDPG) algorithm, which employs an Actor-Critic architecture to handle continuous action spaces in MuJoCo. Fujimoto et al. [24] proposed an improved Twin Delayed Deep Deterministic policy gradient (TD3) algorithm, which mitigates overestimation bias using dual Critic networks. Schulman et al. [25] developed the Proximal Policy Optimization (PPO) algorithm, which introduces a trust-region constraint to enhance robustness and generalization across multiple tasks such as Hopper-v2 and Walker2d. Furthermore, the Soft Actor-Critic (SAC) algorithm further enhances robustness through a maximum entropy objective [26]. However, these algorithms primarily rely on densely connected artificial neural networks (ANNs), whose inherent high computational redundancy and continuous energy consumption pose severe bottlenecks for deployment on resource-constrained edge devices—failing to satisfy the stringent real-time responsiveness and low-power demands of robotic dynamic control in embedded scenarios.

To address this critical gap, spiking RL integrates the temporal dynamics of spiking neural networks (SNNs) with the decision-making mechanisms of RL. Owing to its intrinsic event-driven sparsity, spiking RL offers remarkable energy efficiency and has emerged as a viable paradigm for low-power real-time control [27]-[30]. Notable advancements have been achieved in this domain: Tang et al. [31] first combined Deep Q-Network (DQN) with SNN, achieving a 75-fold improvement in energy efficiency over conventional DDPG methods in mobile robot navigation tasks. Subsequently, the same group proposed PopSAN [32], which encodes observations and actions as neural population activity, attaining a 140-fold improvement in energy efficiency in robotic continuous control tasks compared to Jetson TX2. Patel et al. [33] verified the feasibility of spiking RL on Pendulum-v1 task in MuJoCo via an ANN-to-SNN conversion strategy. Liu et al. [34] further proposed a direct training method based on surrogate gradients, which substantially enhances the performance of spiking RL in continuous control tasks. Additionally, spiking RL has been successfully applied to gait generation in multi-legged robots [35], lane keeping [36], target tracking [37]-[38], and target navigation [39], among others [40].

Although spiking RL demonstrates remarkable energy efficiency, its deployment on conventional electronic computing architectures remains significant challenges. Matrix computations in fully connected layers are inherently limited by RC delays and the von Neumann bottleneck, yielding single-operation latencies on the order of tens of nanoseconds [41]-[45]. In continuous control tasks such as those in MuJoCo, these delays can readily introduce phase lags within the control loop, resulting in gait instability [19], [46]. In contrast, photonic computing, characterized by intrinsic parallel transmission, low propagation loss, and picosecond-scale latency, has emerged as a highly promising solution to the computational bottlenecks inherent to electronic systems. In recent years, it has evolved into a multifaceted technological ecosystem for neural network acceleration, encompassing hardware implementations based on Mach-Zehnder interferometer (MZI) [47]-[53], micro-ring resonator (MRR) [54]-[60], and phase-change material (PCM) [61]-[63]. Among these, programmable MZI meshes can execute matrix-vector multiplication (MVM) by modulating optical phases. The associated computation latency depends solely on light propagation time, offering excellent scalability, and these meshes have demonstrated substantial acceleration in static data processing tasks such as handwritten digit recognition [64]-[67].

While photonic computing and spiking RL each offer unique advantages, their deep integration within robotic continuous control tasks remains nascent. Most existing photonic computing architectures are tailored for densely connected ANNs and remain incompatible with the sparsity and discrete activation characteristics of spiking signals [68]-[71]. Prior efforts have been restricted to simple control environments such as Pendulum and CartPole [72]-[73], without a comprehensive framework encompassing weight mapping, in-situ hardware training, and closed-loop validation. Furthermore, most photonic SNN research has focused on static tasks (e.g., image or speech processing [74]-[80]), leaving dynamic control applications largely unexplored. These gaps underscore the urgent need for dedicated frameworks that bridge photonic hardware and dynamic robotic control.

In this work, we systematically investigated software-hardware collaborative design of MZI-based photonic computing within spiking RL control to establish an optical-electronic hybrid computing architecture tailored for robotic continuous control, providing a novel pathway toward high real-time, low-power embedded intelligent control. The core contributions are as follows: (1) we designed and fabricated a simplified 16×16 MZI mesh chip to realize photonic linear matrix computation; (2) we proposed an optical-electronic hybrid computing architecture combining optical-domain linear transformation with electrical-domain nonlinear activation, wherein the spiking Actor network of the TD3 algorithm is deployed onto the programmable MZI mesh; (3) we designed a weight reconstruction and calibration strategy based on stochastic parallel gradient descent (SPGD), and conducted multi-level simulation verification on Pendulum-v1 and HalfCheetah-v2 tasks; and (4) we constructed a hardware experimental system integrating a field-programmable gate array (FPGA) interface, MZI computing core, and high-speed optical-electronic conversion to achieve full optical-electronic closed-loop control from state input to action output in the HalfCheetah-v2 task.

In experimental validation, the MZI neuromorphic chip achieved an energy efficiency of 1.39 TOPS/W and a computational density of 0.13 TOPS/mm² for linear computation, with an end-to-end latency of 120 ps in the photonic SNN layer. In the HalfCheetah-v2 task, the hardware inference outputs deviated by less than 2.2% from those of the pure software implementation, while the hardware-software collaborative computing achieved a stable reward of $5831\pm495$, reducing the convergence steps by 23.33% compared with the fully electronic baseline. Furthermore, the proposed optical-electronic hybrid acceleration scheme can be readily extended to other continuous control tasks in MuJoCo, such as Pendulum-v1, providing an energy-efficient and high-performance computational foundation for applications including real-time industrial robotic control and bionic robot gait planning.

## 2. Concept and architecture

### 2.1 Environment and tasks

MuJoCo (Multi-Joint dynamics with Contact), developed by Todorov's team at the University of Washington, is a high-precision physics engine for multibody systems and a 3D simulation platform widely adopted in robotics, biomechanics, and computer graphics [81]. Compared with OpenAI Gym [82], MuJoCo, when integrated with DeepMind's open-source Control Suite [83], is particularly well suited for reinforcement learning tasks involving high-dimensional continuous action spaces. In robotic motion simulation, MuJoCo provides both higher computational efficiency and improved simulation fidelity through fine-grained physical modeling, thereby preventing performance misjudgments that may arise from low-fidelity approximations. Currently, the platform is extensively employed for benchmark tasks such as Pendulum-v1 (single pendulum balancing), HalfCheetah-v2 (bipedal robot running), and Hopper-v2 (single-leg hopping), and has become a standardized tool for evaluating the performance of continuous control algorithms.

To validate our approach, we selected two representative continuous control tasks from MuJoCo: Pendulum-v1 and HalfCheetah-v2. Pendulum-v1 serves as a canonical balancing task, where the goal is to keep an inverted pendulum upright by applying continuous torque to the pendulum rod. Its reward function, defined in Equation (1), penalizes angular deviation, angular velocity, and control energy to achieve stable balance. HalfCheetah-v2, in contrast, represents a high-dimensional locomotion task, requiring a cheetah-like bipedal robot to learn coordinated joint actuation for fast and stable forward motion. The corresponding reward function in Equation (2) promotes forward velocity while discouraging excessive energy consumption. The detailed parameters of the state space, action space, and other environment settings for both tasks are summarized in Table 1.

$$r = -(\theta^2) + 0.1 \cdot \dot{\theta}^2 + 0.001 \cdot u^2 \tag{1}$$

$$r = forward\_reward - 0.1 \cdot action\_cost = 1.0 \cdot \frac{\Delta x}{\Delta t} - 0.1 \cdot \sum_{i=1}^{6} action_i^2 \tag{2}$$

where, $\theta$ denotes the angular deviation of the pendulum from the vertical position, $\dot{\theta}$ represents the angular velocity of the pendulum, and $u$ denotes the control torque applied to it. $forward\_reward$ refers to the reward associated with forward motion, while $action\_cost$ represents the control cost. $\Delta x$ indicates the incremental displacement of the robot's torso along the x-axis per step, and $\Delta t$ denotes the fixed time step for each action. $action_i$ represents the control torque applied to the $i$-th joint.

Table 1 Environmental parameters

| Category | Parameter | Value |
| --- | --- | --- |
| Pendulum-v1 | Cos(θ) | [-1, 1] |
| | Sin(θ) | [-1, 1] |
| | Initial θ | [-π, π] rad |
| | Angular Velocity θ_dot | [-8, 8] rad/s |
| | Initial θ_dot | [-1, 1] rad/s |
| | Torque range | [-2, 2] N•m |
| | Pendulum Length | 1.0 m |
| | Pendulum Mass | 1.0 kg |
| | Gravity | 9.8 m/s² |
| | State dimension | 3 |
| | Action dimension | 1 |
| HalfCheetah-v2 | Trunk z-coordinate | (-Inf, Inf) m |
| | Trunk pitch angle | (-Inf, Inf) rad |
| | 6 joint angles | (-Inf, Inf) rad |
| | Trunk pitch angular velocity | (-Inf, Inf) rad/s |
| | 6 joint angular velocities | (-Inf, Inf) rad/s |
| | Trunk x-coordinate velocity | (-Inf, Inf) m/s |
| | Trunk z-coordinate velocity | (-Inf, Inf) m/s |
| | 6 joint torques | [-1, 1] N•m |
| | Fixed time step per actions step | 0.05 s |
| | Ground friction coefficient | 0.8 |
| | State dimension | 17 |
| | Action dimension | 6 |

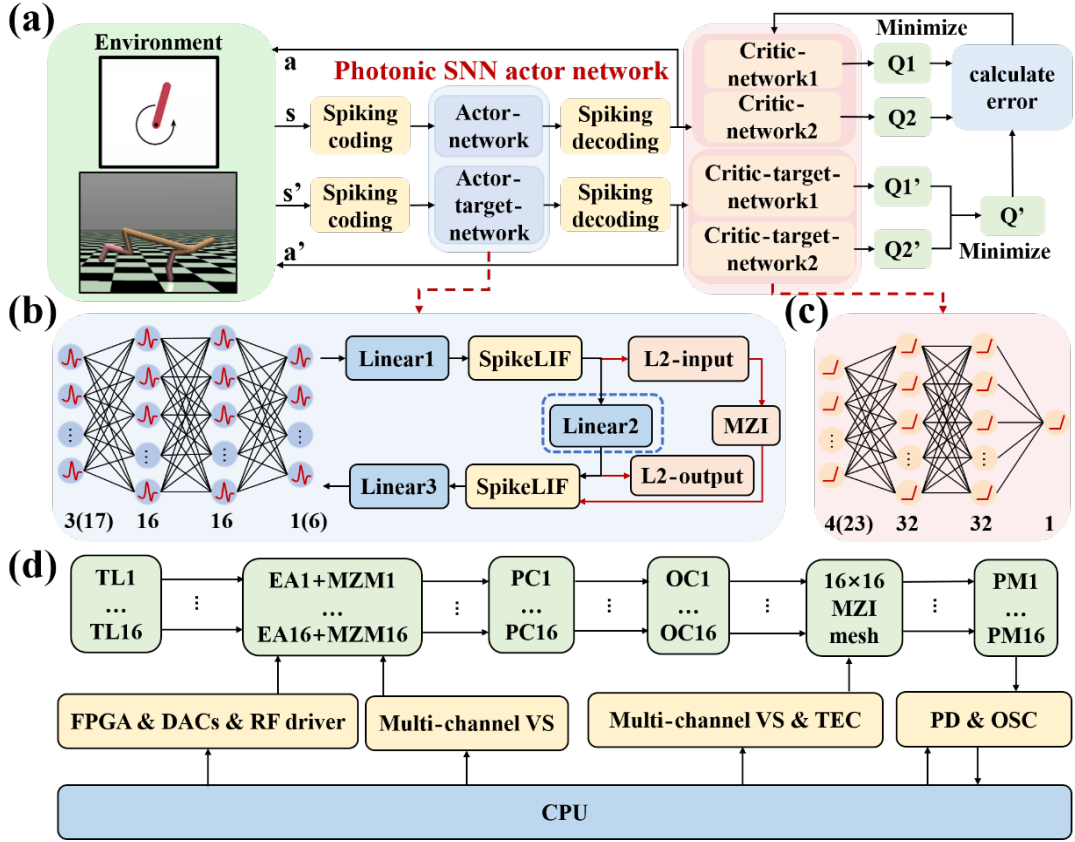

Figure 1 Overall Framework. (a) Overall architecture of the photonic spiking RL framework based on the TD3 algorithm; (b) structure of the Actor network; (c) structure of the Critic network; (d) schematic diagram of the experimental setup.

## 2.2 Spiking Reinforcement Learning-TD3 integrated framework

We proposed a spiking TD3 algorithm that extends the classical TD3 framework for neuromorphic and optical-electronic collaborative computing. The primary enhancements involve converting the network architecture into a spiking form and adapting its dimensionality to the specific requirements of each task. As illustrated in Fig. 1(a), the algorithm preserves the overall dual-Critic and dual-target Critic architecture of TD3. The value function is optimized by minimizing both the Q-value and the target Q-value errors, whereas the Actor network is redesigned as a SNN employing the leaky integrate-and-fire (LIF) model. For the Pendulum-v1 and HalfCheetah-v2 tasks, the Actor network is configured as $3 \times 16 \times 16 \times 1$ and $17 \times 16 \times 16 \times 6$, respectively, while the Critic networks are structured as $4 \times 32 \times 32 \times 1$ and $23 \times 32 \times 32 \times 1$, as shown in Fig. 1(b)-(c). The temporal resolution, represented by the spiking time step T, is set to 1 and 4 for the two tasks, respectively. During training, the agent continuously interacts with the MuJoCo environment: the SNN Actor network generates task-specific actions, executes them, and stores the resulting transitions in a replay buffer. Once the buffer accumulates sufficient samples, the algorithm randomly samples mini-batches for network updates. The Critic networks refine value estimation by minimizing the Q-value loss, while the SNN Actor network performs delayed policy updates at fixed intervals. The target Actor and Critic networks are synchronized with the current networks through a soft-update mechanism. The algorithmic process is summarized in Table 2. Within the SNN Actor network, temporal computation is unrolled along the time dimension T: continuous state inputs are encoded into spike sequences, processed through multiple LIF-Spike activation layers, temporally averaged, and then mapped onto the action space. This mechanism bridges the transformation from discrete spiking signals to continuous control actions.

Table 2 Algorithm flowchart

| **Spiking TD3 Algorithm Process** |
| --- |
| 1. Initialize SNN Actor network $\pi_\phi$, ANN Critic networks $Q_{\theta_1}$, $Q_{\theta_1}$, and target networks $\pi_{\phi'}$, $Q_{\theta'_1}$, $Q_{\theta'_2}$ |
| 2. Configure optimizers and replay buffer $\mathcal{B}$ |
| 3. Set random seeds and training parameters |
| 4. **for** $t = 1$ **to** $T$ **do:** |
| 5.     Generate action $a \sim \pi_\phi(s) + \mathcal{N}(0, \sigma)$; obtain the next state $s'$ and observe reward $r$ |
| 6.     Store transition tuple $<s, a, r, s'>$ in $\mathcal{B}$ |
| 7.     **if** $\|\mathcal{B}\|$ > buffer size: |
| 8.         Randomly sample a mini-batch from data from $\mathcal{B}$ |
| 9.         Update the Critics: $a' \sim \pi_{\phi'}(s') + \epsilon, \epsilon \sim clip(\mathcal{N}(0, \tilde{\sigma}), -c, c)$ |
| 10.         $y \leftarrow r + \gamma \, min\{Q_{\theta'_1}(s', a'), Q_{\theta'_2}(s', a')\}$ |
| 11.         $\theta_i \leftarrow argmin_{\theta_i} N^{-1} \sum (Q_{\theta_i}(s, a) - y)^2$ |
| 12.         **if** $t \bmod d = 0$ **then:** |
| 13.             Update $\phi$ by the policy gradient: $\nabla_\phi J(\phi) = N^{-1} \sum \nabla_a Q_{\theta_1}(s, a)|_{a=\pi_\phi(s)} \nabla_\phi \pi_\phi(s)$ |
| 14.             Update target networks: $\theta'_i \leftarrow \tau \theta_i + (1 - \tau)\theta'_i$, $\phi' \leftarrow \tau \phi + (1 - \tau)\phi'$ |
| 15.         **end if** |
| 16.     **end if** |
| 17. **end for** |
| 18. Save the final trained model |

## 2.3 Software-hardware co-training and inference pipeline

To validate the integration effectiveness of the proposed spiking TD3 algorithm with photonic hardware, a systematic end-to-end verification pipeline is established, spanning from algorithm design to hardware deployment. We design and implement a three-layer experimental verification framework encompassing the entire workflow from algorithm preparation and hardware mapping to performance evaluation. The schematic diagram of the photonic hardware implementation is shown in Fig. 1(d), and the detailed verification procedure is as follows:

(1) Software pre-training and weights extraction: The spiking TD3 algorithm is first trained independently in MuJoCo, where the Actor network adopts an SNN architecture and the Critic networks employ conventional ANNs. The SNN is trained from scratch using the surrogate gradient method with T temporal steps. Upon convergence, the target weights matrix of the second linear layer ($L_2$) in the Actor network, together with its corresponding input-output test dataset, is extracted for hardware mapping. This stage is entirely conducted in the digital domain and provides a reliable algorithmic baseline for subsequent hardware verification.

(2) Hardware mapping and in-situ calibration: The target matrix obtained in step (1) is mapped onto the MZI-based photonic chip. A stochastic parallel gradient descent (SPGD) algorithm is employed to iteratively optimize the driving voltages of each thermo-optic phase shifter in the MZI array, thereby minimizing the discrepancy between the experimentally measured transmission matrix and the desired target matrix under specified optical inputs. This in-situ calibration compensates for fabrication imperfections and system noise, enabling high-precision optical computation.

(3) Offline hardware inference and comparative validation: Following calibration, offline hardware inference experiments are performed. The input vectors from the test dataset are converted into optical signals via the FPGA and modulator array, then injected into the configured MZI chip. The optical outputs are collected and converted into electrical signals to obtain the corresponding hardware responses. Finally, these hardware responses are compared one-to-one with the software-derived baseline outputs, and the inference accuracy is computed to quantitatively evaluate the performance of the photonic computing layer.

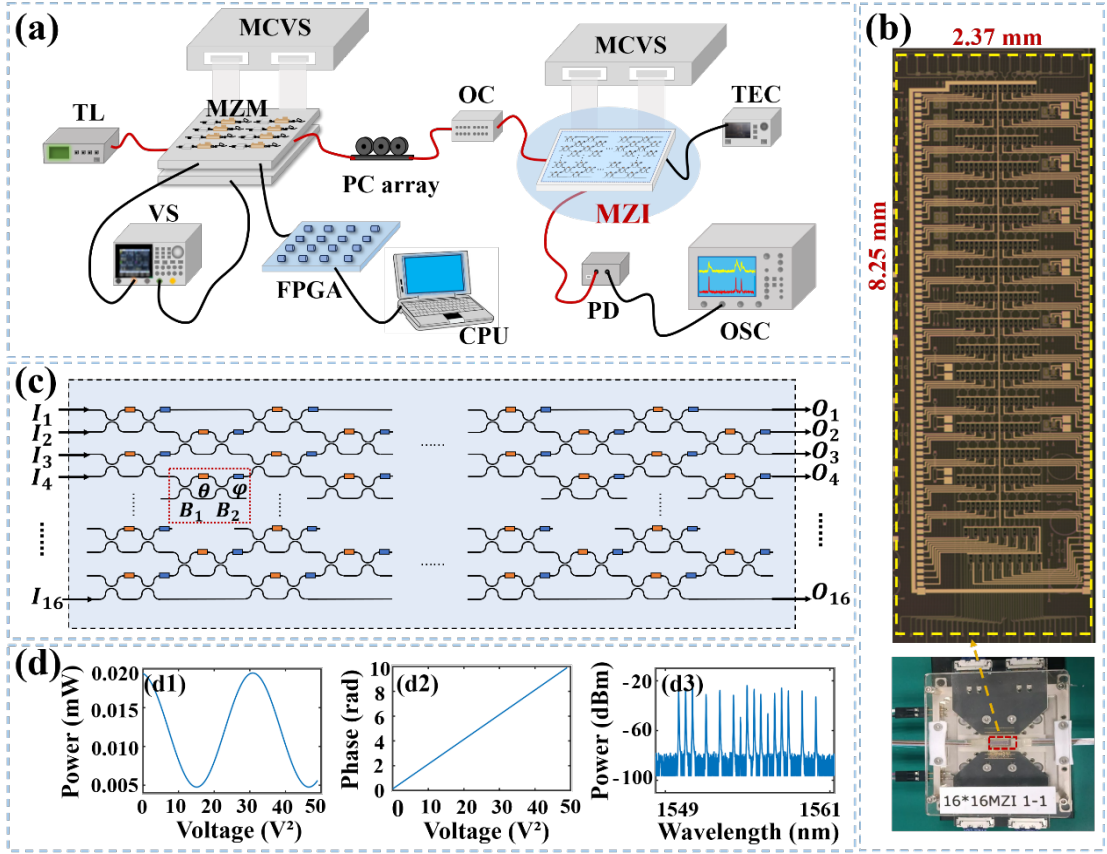

Figure 2 Schematic of the experimental setup for testing the photonic neuromorphic computing chip module and its characteristics. (a) Experimental setup diagram; (b) unpackaged and packaged views of the photonic chip; (c) the 16×16 MZI mesh; (d) basic performance testing of the MZI chip.

## 3 Result

### 3.1 Photonic neuromorphic chip

We designed and fabricated a simplified 16×16 MZI mesh chip on a silicon-on-insulator (SOI) platform. Unlike the conventional unitary-diagonal-unitary network derived from singular value decomposition (SVD), which imposes stringent structural constraints on scalability and control flexibility. To address these limitations, we proposed an innovative hybrid topology that combines an (N+1)×(N+1) unitary MZI mesh with an N×N diagonal matrix (N=16), effectively approximating the functionality of an N×N synaptic weight matrix. The resultant 16×16 MZI mesh structure is illustrated in Fig. 2(c). This simplified design substantially reduces the number of phase shifters and the associated control overhead. As shown in Fig. 2(b), the fabricated chip exhibits compact dimensions of 2.37 mm×8.25 mm, corresponding to a total footprint of 19.55 mm² . It integrates 152 thermo-optic phase shifters based on titanium nitride (TiN) heaters, enabling parallel optical-domain computation for 16×16 real-valued matrix operations. Experimental characterization demonstrates a linear MVM throughput of approximately 2.5 tera-operations per second (TOPS), with a total power consumption of about 1.8 W.

The measured computation latency is as low as 120 ps, corresponding to an energy efficiency of 1.39 TOPS/W and a computational density of 0.13 TOPS/mm² .

We employed the self-designed MZI chip to implement the optical-domain MVM for the 16×16 $L_2$ layer within the Actor network of the spiking TD3 algorithm. Each MZI unit modulates the phase of light through the thermo-optic effect and utilizes optical interference to control signal intensity, thereby accomplishing the linear transformation between the weight matrix and input vectors. By precisely adjusting the drive voltages of 152 phase shifters, the system efficiently performs MVM in the optical domain. The optical linear computation process combines nanosecond-level low latency with inherent parallelism, effectively overcoming the bandwidth and power consumption bottlenecks inherent of conventional electronic architectures for matrix operations.

*3.2 Experimental system setup*

The experimental setup integrates a comprehensive optical signal chain encompassing signal generation, modulation, transmission, processing, and detection, as shown in Fig. 2(a). A tunable laser (TL) generates a continuous-wave optical signal, which is subsequently modulated into spiking-sequence-matched optical signals using an electrical amplifier (EA) and a Mach-Zehnder modulator (MZM). A polarization controller array (PC array) adjusts and maintains the polarization state before the optical signals are injected into the simplified MZI mesh chip. A multi-channel voltage source (MCVS) provides precise bias voltages to tune the MZI phase shifters, while a thermoelectric cooler (TEC) stabilizes the chip temperature at 25°C to minimize thermal fluctuations that could disturb phase stability. The optical outputs from the MZI array are converted into electrical signals via a photodetector (PD) and subsequently captured and analyzed by an oscilloscope (OSC), thereby enabling electrical-domain readout of the optical-domain linear computation results. A FPGA and a central processing unit (CPU) operate cooperatively to manage digital-to-analog signal conversion, control command delivery, and data post-processing, ensuring efficient synchronization across the optical-electronic computation chain.

*3.3 Optical characterization and fundamental testing*

To verify the modulation capability of the MZI unit for optical signals, we conducted a fundamental optical characterization experiment, as illustrated in Fig. 2(d). We focused on three key aspects, interference response, phase linearity, and wavelength stability, and systematically evaluated the feasibility of the MZI chip for optical-domain linear computation. First, we measured the relationship between the output optical power of the MZI unit and the control voltage applied to the phase shifters. The result shows that the optical power varies periodically in a sinusoidal manner with voltage, consistent with the theoretical characteristics of MZI. Second, we examined the correspondence between the optical phase of the MZI unit and the control voltage applied to its phase shifter. The result demonstrates a clear linear relationship between the two, with a linear fitting coefficient of $R^2 > 0.99$. Finally, we characterized the optical response of the MZI system within the target wavelength range of 1549-1561 nm. The measured wavelength spectrum exhibits uniformly distributed peaks without noticeable fluctuations or attenuation, indicating a stable optical response across the operating wavelength range. Through these experiments, we validated the feasibility and stability of the MZI chip and its integrated hardware system for optical-domain linear computation.

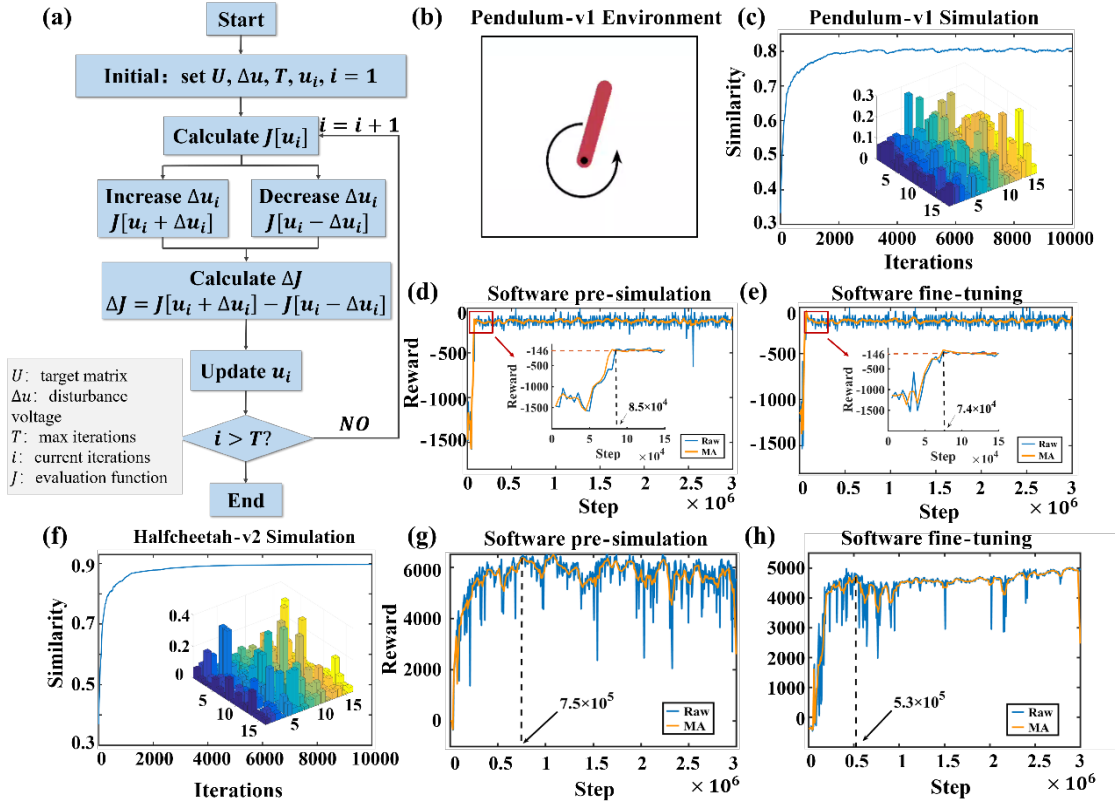

Figure 3 Experimental simulation. (a) Stochastic parallel gradient descent algorithm flow; (b) test environment of Pendulum-v1; (c)-(e) simulation training for Pendulum-v1; (c) MZI training; (d) software pre-simulation reward; (e) post-simulation software fine-tuning reward; (f)-(h) Halfcheetah-v2 simulation training; (f) MZI training; (g) software pre-simulation reward; (h) post-simulation software fine-tuning reward.

*3.4 Simulation of the Pendulum-v1 and HalfCheetah-v2 tasks*

To evaluate the suitability and feasibility of photonic computing hardware for spiking RL tasks, we developed a high-fidelity optical-electronic hybrid simulation platform for two continuous control tasks, Pendulum-v1 and HalfCheetah-v2 in MuJoCo. This platform enables collaborative simulation through the integration of physical models and algorithmic modules. Built upon MATLAB simulation tools, this platform integrates a high-precision physical model of MZI chips, enabling accurate simulation of the optical transmission characteristics and weights configuration processes of MZI arrays. At the device level, the transmission matrix method was employed to model MZI units physically, ensuring that their optical responses align with actual device characteristics. At the algorithm level, a complete SPGD optimization simulation model was constructed (Fig. 3(a)), consisting of three core modules: random perturbation generation, gradient estimation, and voltage update. During simulation, for the Pendulum-v1 task (time step T=1) and the HalfCheetah-v2 task (time step T=4), the $L_2$ layer weights of the Actor network trained to convergence by the spiking RL algorithm were used as the target matrices. Through iterative optimization using the SPGD model, we generated the voltage configuration table for the MZI phase shifters and the corresponding hardware equivalent weights matrices, thereby completing the simulation of the mapping from digital-domain weights to optical-domain hardware.

The simulation results demonstrate that the SPGD algorithm effectively maps weights matrices from the digital-domain to the optical-domain. For the Pendulum-v1 task, the cosine similarity between trained output weights and target weights rapidly increased with iterations, converging to 0.81 after approximately 2000 iterations. For the more complex HalfCheetah-v2 task, the similarity ultimately stabilizes at 0.89. Regarding task performance, after optical-domain mapping, the reward curve for Pendulum-v1 converged faster with reduced steady-state fluctuations. It achieved a stable reward value (-146) comparable to that of the pure algorithm in only $7.4 \times 10^4$ steps. In the HalfCheetah-v2 task, the optical-electronic hybrid architecture effectively mitigated the pronounced oscillations observed during training with the original algorithm. The reward curve exhibited a smoother upward trend, with significantly reduced performance fluctuations during the steady phase. Compared with the pure algorithm, the convergence steps were reduced by 29.33%.

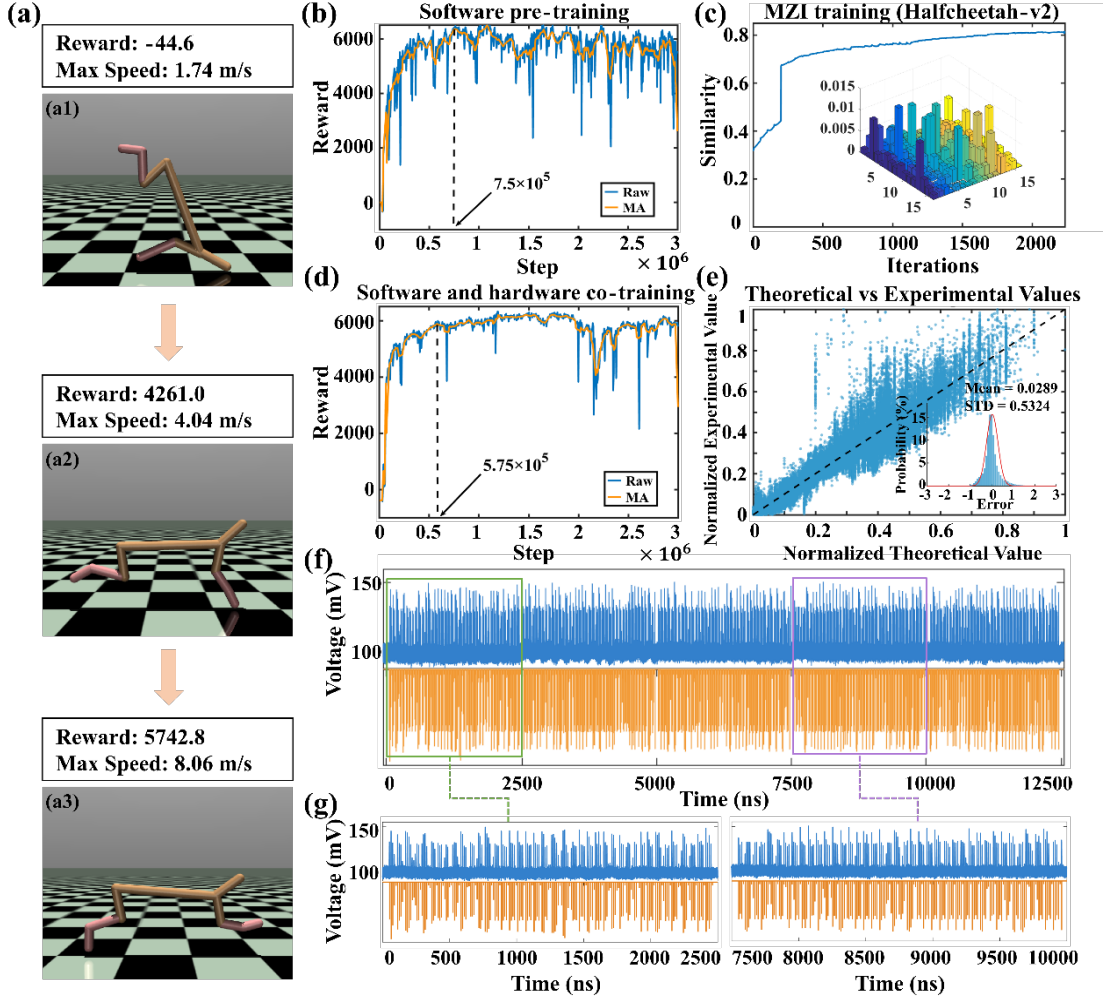

Figure 4 Training and inference results for the HalfCheetah-v2 task. (a) Motion states of HalfCheetah-v2 under three reward values; (b) software pre-training reward; (c) MZI training; (d) software-hardware co-training fine-tuned training reward; (e) comparison of experimental outputs with theoretical values; (f)-(g) experimentally measured linear outputs.

### 3.5 HalfCheetah-v2 continuous control tasks based on photonic neuromorphic chip

To validate the performance of the optical-electronic hybrid architecture in RL tasks, we conducted experiments on the HalfCheetah-v2 continuous control task. These experiments included software pre-training, MZI hardware weights mapping, and collaborative training between software and hardware. The experimental results are shown in Fig. 4.

Figure 4(a) shows the test environment interface for the HalfCheetah-v2 continuous control task. The movement state and velocity of the HalfCheetah-v2 agent vary with different reward values. Figure 4(b) presents the reward curves from software pre-training, where the blue curve represents the raw reward and the orange curve shows the reward after moving average (MA) smoothing. The curves indicate that the reward values increase rapidly from initial lows, gradually stabilizing around $7.5 \times 10^5$ steps and finally converging near 6000. However, the raw reward exhibits significant fluctuations. Subsequently, MZI weights mapping was conducted using the SPGD algorithm. Figure 4(c) shows the SPGD training results, including the similarity curve across iterations and the 3D similarity distribution. As iterations increase, the cosine similarity between hardware-equivalent and target weights increases rapidly from an initial value of approximately 0.3. After about 1500 iterations, it approaches 0.81 and remains stable thereafter. Figure 4(d) presents the reward curves for hardware-software collaborative training, also shown as RAW and MA. Compared with pure software training, the reward value increases substantially faster under hardware-software collaborative training, achieving convergence of $5831 \pm 495$ earlier. After convergence, the amplitude of fluctuations decreases markedly, and the MA curve becomes smoother. These results indicate that the hardware-software collaborative architecture effectively enhances the stability and convergence rate of policy training.

In the photonic hardware inference experiment, we considered 1000 state-action pair samples, deploying a $16 \times 16$ critical layer on the photonic neuromorphic chip. Figure 4(e) shows the normalized theoretical and experimental results, along with their error distributions for 16 channels (time step T=4) using 1000 state-action pairs and a total transmitted data volume of 64000. The error probability distribution curve indicates that errors are predominantly concentrated around 0 and exhibit features consistent with a normal distribution. The experimental results show strong agreement with theoretical predictions, indicating that the system maintains excellent stability and accuracy during multi-channel data transmission. Figure 4(f)-(g) compare the target and experimentally measured outputs for 1000 state-action pairs transmitted by Channel 1 at T=1, including a locally magnified view. The experimental (blue curve) and target outputs (orange curve) exhibit highly consistent overall trends, demonstrating strong temporal alignment with only minor deviations.

## 4 Discussion

Here, we further investigate the performance of MZI chips at larger scales. In the algorithm design, SNN Actor networks with hidden layers of $32 \times 32$, $64 \times 64$, and $128 \times 128$ were also considered. The corresponding Critic networks were configured with one additional hidden layer compared with the Actor networks. The results, shown in Fig. 5(b), demonstrate that larger networks achieve higher reward values than those with $16 \times 16$ hidden layers. Furthermore, we compared the reward performance across different time step T for this task, specifically T=1, T=2, and T=4. As illustrated in Fig. 5(a), convergence becomes faster as the time step T increases. In Table 3, we compare the proposed scheme with existing approaches. Our work achieves optical-electronic co-inference on a photonic chip using a small $16 \times 16$ hidden layer, yielding stable reward values.

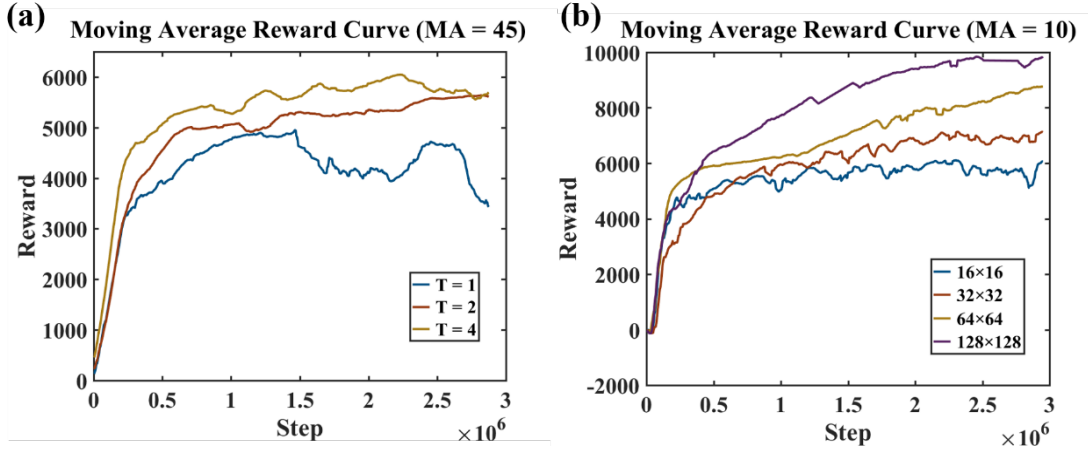

Figure 5 The reward for a Spiking TD3 for HalfCheetah-v2 task. (a) The reward with T=1, 2, 4; (b) the reward with hidden layer size of 16×16, 32×32, 64×64, and 128×128.

Table 3 Performance comparison

| Reference | [24] | [40] | [70] | [84] | [85] | **Our work** |
|---|---|---|---|---|---|---|
| HalfCheetah version | v1 | v3 | v2 | v3 | v2 | **v2** |
| Method | TD3 | SNN-TD3 | NNTD3 | Q-MLP TD3 | TD3 | **Photonic SNN TD3** |
| Reward | 9636.95±859.065 | 11000±751 | 8341.74 | 10295.54 | 9842.387±1013.92 | **5831±495** |
| Hidden layer size | 400×300 | 256×256 | 32×32 | 192×192 | 400×300 | **16×16** |
| For SNN | × | ✓ | × | × | × | ✓ |

## 5 Conclusion

For robotic continuous control tasks, addressing the conflict between high-dimensional state real-time response and low power consumption exemplified by HalfCheetah-v2, this work proposes and experimentally validates a novel architecture for photonic spiking RL. This architecture achieves a deep integration of photonic computing and spiking RL, thereby bridging the gap in photonic computing applications for robotic dynamic control. The proposed framework follows a hybrid paradigm that performs linear transformations in the photonic domain and nonlinear activations in the electrical domain. By deploying the 16×16 weights of Actor network onto the photonic MZI chip, it achieves high energy efficiency (1.39 TOPS/W), high computational density (0.13 TOPS/mm$^2$), and low latency (120 ps). Sparse computation is realized in the electrical-domain through LIF SNN. After SPGD calibration, the MZI hardware weights achieved 81% cosine similarity to the target weights. On the HalfCheetah-v2 task, hardware-software collaborative computing achieved a reward value of 5831±495, with less than 2.2% action deviation compared with pure software inference. The convergence steps were significantly reduced, and the proposed architecture can be generalized to other robotic continuous control tasks.

## Data availability

The data that support the findings of this study are available from the corresponding author upon reasonable request.

## Code availability
The source codes of this study are available in the repository of the project.

## Acknowledgements

This work was supported by National Natural Science Foundation of China (No. 62535015); The Fundamental Research Funds for the Central Universities (QTZX23041).

## Competing interests
The authors declare no competing interests.